\documentclass[conference,compsoc]{IEEEtran}
\ifCLASSOPTIONcompsoc
  \usepackage[nocompress]{cite}
\else
  \usepackage{cite}
\fi
\ifCLASSINFOpdf
\else
\fi

\usepackage{times}
\usepackage{epsfig}
\usepackage{graphicx}
\usepackage{amsmath}
\usepackage{amssymb}
\usepackage{threeparttable}
\usepackage{multirow}
\usepackage{booktabs}
\usepackage{subfigure}
\usepackage{cite}
\usepackage{graphicx}
\usepackage{float}
\usepackage{amsmath,amssymb}
\usepackage{color}
\usepackage[colorlinks]{hyperref}

\usepackage{geometry}
\begin{document}
%
\title{PRAFlow\_RVC: Pyramid Recurrent All-Pairs Field Transforms for Optical Flow Estimation in Robust Vision Challenge 2020}

\author{\IEEEauthorblockN{Zhexiong Wan\IEEEauthorrefmark{1},
Yuxin Mao\IEEEauthorrefmark{2},
Yuchao Dai\IEEEauthorrefmark{3}}
\IEEEauthorblockA{School of Electronics and Information, 
Northwestern Polytechnical University, Xi’an, China.}
\IEEEauthorblockA{\IEEEauthorrefmark{1} wanzhexiong@mail.nwpu.edu.cn \IEEEauthorrefmark{2} maoyuxin@mail.nwpu.edu.cn \IEEEauthorrefmark{3} daiyuchao@nwpu.edu.cn (Corresponding author)}}


\maketitle

\pagestyle{plain}
\pagenumbering{arabic}
\thispagestyle{plain}

\begin{abstract}

Optical flow estimation is an important computer vision task, which aims at estimating the dense correspondences between two frames. 
RAFT (Recurrent All Pairs Field Transforms)\cite{teed2020raft} currently represents the state-of-the-art in optical flow estimation. It has excellent generalization ability and has obtained outstanding results across several benchmarks. To further improve the robustness and achieve accurate optical flow estimation, we present PRAFlow (Pyramid Recurrent All-Pairs Flow), which builds upon the pyramid network structure. Due to computational limitation, our proposed network structure only uses two pyramid layers. At each layer, the RAFT unit is used to estimate the optical flow at the current resolution. Our model was trained on several simulate and real-image datasets, submitted to multiple leaderboards using the same model and parameters, and won the 2nd place in the optical flow task of ECCV 2020 workshop: Robust Vision Challenge\footnote{http://www.robustvision.net/leaderboard.php?benchmark=flow}.

\end{abstract}


%
\IEEEpeerreviewmaketitle

\section{Introduction}

Optical flow estimation is a typical task in computer vision which aims to build the relation between frames by exploiting the photometric consistency. In its most standard setup, optical flow is estimated from the input two frames and represents the displacement of the corresponding pixels from the first frame to the next frame. With the application of deep learning in computer vision tasks, optical flow estimation algorithms have also been greatly advanced. 


As an influential algorithm in learning optical flow, PWC-Net\cite{sun2018pwc} extracts features through pyramid processing and builds a cost volume at each level from warped feature to iteratively refine the estimated flow. Very recently, Recurrent All-Pairs Field Transforms (RAFT)\cite{teed2020raft} builds multi-scale 4D correlation volumes for all pairs of pixels, and iteratively updates a flow field through a gated recurrent unit (GRU) that performs lookups on the correlation volumes. Through the above improvements, RAFT achieves the state-of-the-art results across many optical flow benchmarks.

Motivated by the RAFT architecture and the classic pyramid coarse-to-fine structure, we propose Pyramid Recurrent All-Pairs Field Transforms (PRAFlow) to achieve more accurate high-resolution optical flow estimation. 
In our experiment, due to the limit of computation, we only use two pyramid layers. The weight shared RAFT unit is used to estimate the 1/8 and 1/4 resolution flow respectively in the two layers. Finally, we use the 4x convex upsample on 1/4 resolution predict flow to get the final full resolution flow similar to RAFT.

\section{Approach}
Our model is improved upon RAFT\cite{teed2020raft}, 
which consists of three main components: (1) a feature encoder that extracts per pixel features; (2) a multi-scale 4D corrleation volumes for all pairs of pixels; (3) a recurrent unit that performs lookups on the corrleation volumes. Inspired by RAFT and the pyramid structure in PWC-Net, we consider that estimating optical flow from a coarse-to-fine manner will further improve the final result. 

In our model, the flow is initialized to zero at a resolution of 1/8 image size and then input to RAFT unit, and get the fine optical flow at 1/8 resolution. After that, we do a 2x upsample to get initial coarse flow at 1/4 resolution and perform the same recurrent operation as the previous layer, and do 4x convex sample like RAFT\cite{kondermann2016hci} to get final fine optical flow at full resolution. For each stage in our model, we set the iteration number of RAFT unit to 12 for each benchmarks during both training and evaluation process. 

\begin{figure*}[t] 
\centering 
\includegraphics[width=0.95\textwidth]{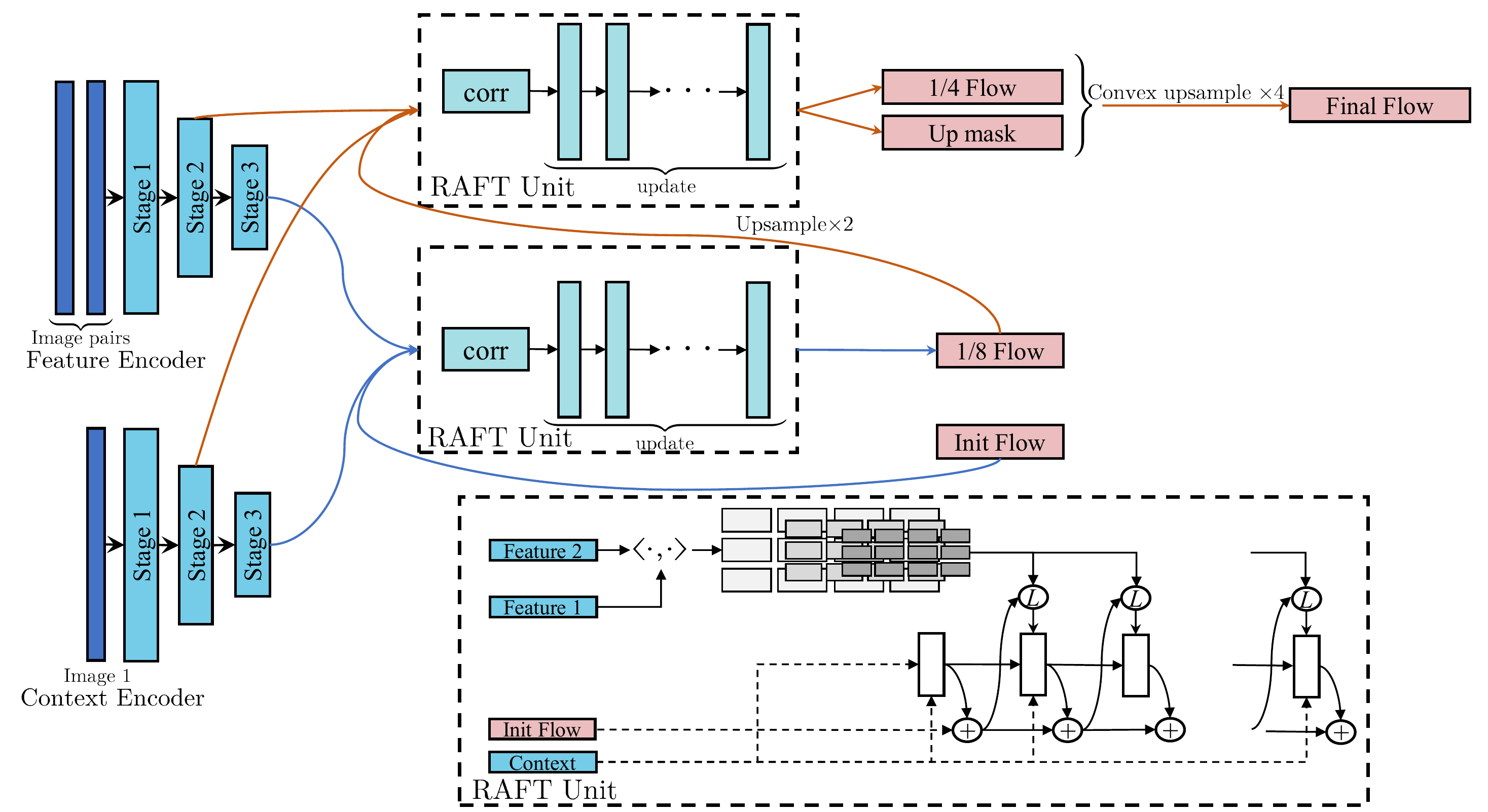} 
\caption{\textbf{Our Pyramid Recurrent All-Pairs Flow (PRAFlow) structure}, which consists of 2 main components: (1) a feature encoder for extracting pyramid features of input image pairs, and a context encoder for extracting pyramid features of input image 1. (2) two RAFT uint in different pyramid stages to update optical flow by using the current estimate to look up the set of correlation volumes in a coarse-to-fine manner.} 
\label{fig-PRAFlow}
\end{figure*}
\section{Experiments}
\subsection{Implementation Details}
Our PRAFlow is implemented in PyTorch\cite{paszke2019pytorch}. All modules are initialized from scratch with random weights. We submit our method to optical flow task of the ECCV 2020 Robust Vision Challenge with the well-known flow benchmarks, Sintel, KITTI, Middlebury, and VIPER. Following previous works, we first pretrain our model on FlyingChairs\cite{dosovitskiy2015flownet} and FlyingThings3D\cite{mayer2016large} supervisedly, and then followed by joint supervised finetuning that combines all datasets.

\noindent
\textbf{Training Schedule}: We pretrain our model on FlyingChairs\cite{dosovitskiy2015flownet} for 100k iterations with a crop size of 368$\times$496, then train for 200k iterations on FlyingThings3D\cite{mayer2016large} with a crop size of 384$\times$512. To get final submission flow results for Robust Vision Challenge, we fine-tune the model using training data from Sintel\cite{butler2012naturalistic}, KITTI2015\cite{Menze2018JPRS}, HD1K\cite{kondermann2016hci}, and VIPER\cite{richter2017playing}. For the same reason as in PWC-Net\cite{sun2018pwc}, we do not use the Middlebury training data. Due to the different scenarios of different datasets and the different motion scales, we need to balance the number of training samples in each dataset during the joint training process. We use \textit{50*Sintel Clean+50*Sintel Final+500*KITTI+2*HD1K+1*Viper} to balance the number of input samples. We finetune our model with and a crop size of 320$\times$608. All training procedure is conducted by using 4 NVIDIA 2080Ti GPUs, and batch size is 4. 

\noindent
\textbf{Data Augmentation}: For data augmentation, we follow the same method as in RAFT, which includes photometric augmentation, spatial augmentation, and occlusion augmentation. We perform photometric augmentation by randomly perturbing brightness, contrast, saturation, and hue. Spatial augmentation is doing by randomly rescaling and stretching the images. For occlusion augmentation, we also randomly erase rectangular regions in image2 with probability 0.5 to simulate occlusions.

\noindent
\textbf{Special Processing of the Viper Dataset}: 
In the process of processing the viper\cite{richter2017playing} dataset, we found that the ground-truth flow of the car front cover areas may not match the actual situation. At the same time, we find there many pixels in the ground-truth with a motion greater than 300 pixels, which is also seems unreasonable. To reduce their impact on model training, we preprocessed the samples in Viper to remove the pixels with labeled flow greater than 300 pixels, and use only the upper half areas of images without the car front cover (about top 700 pixels wide).

\noindent
\textbf{Major Challenge of Multi-Dataset Joint Training}:
We found that the major challenges in the RVC optical flow challenge was the problem of multi-data set balance. There are differences in the scale of motions, real and synthetic gap between different datasets. For example,
in the process of training the model, we found that too much weight in the KITTI dataset in the training set will make the model perform better for small motions, but this will lead to poor performance of the Sintel dataset, because there are a lot of large motions on the Sintel dataset. In order to achieve better results comprehensively, we use the same weighted ratio for multiple datasets. At the same time, due to the problem of the ground-truth of the Viper data set, we appropriately reduced the weight of the Viper dataset.

\subsection{Main Results}
PRAFlow was submited to the optical flow task of the ECCV 2020 Robust Vision Challenge, and it won the the 2nd place with comparable results on those benchmarks to other optical flow estimation algorithms. Detailed results on the MPI Sintel, KITTI2015, Viper and Middlebury benchmarks as presented in Table \ref{tab:main_results}.
\begin{table*}[t]
    \caption{Validation Results for Robust Vision Challenge}
    \label{tab:main_results}
    \centering
    \begin{threeparttable}
    \begin{tabular}{ccccccccc}
        \toprule
        & \multicolumn{2}{c}{Sintel \textit{Final}} & \multicolumn{2}{c}{Sintel \textit{Clean}} & \multicolumn{2}{c}{KITTI} & Viper & Middlebury\cr
        \cmidrule(lr){2-3}\cmidrule(lr){4-5}\cmidrule(lr){6-7}
        Method & AEPE$\downarrow$ & AEPE$\downarrow$ & AEPE$\downarrow$ & AEPE$\downarrow$ & AEPE$\downarrow$ & 
        F-all$\downarrow$ & m-WAUC$\uparrow$ & avg-Rank$\downarrow$ \cr
               & \textit{train} & \textit{test} & \textit{train} & \textit{test} & \textit{train} & \textit{test} & \textit{test} & \textit{test} \cr
        \midrule
        RAFT-TF\_RVC                        & -    & 3.32 & -    & 1.84 & -    & 5.56\%  & 69.5   & 12.0\cr
        C-RAFT\_RVC                         & -    & 3.80 & -    & 2.29 & -    & 8.75\%  & 61.7   & 78.6\cr
        IRR-PWC\_RVC\cite{hur2019iterative} & -    & 4.80 & -    & 3.79 & -    & 8.38\%  & 68.9   & 105.7\cr
        VCN\_RVC\cite{yang2019volumetric}   & -    & 4.52 & -    & 2.83 & -    & 10.15\% & 60.9   & 44.5\cr
        LSM-Flow\_RVC\cite{tang2020lsm}     & -    & 4.21 & -    & 2.99 & -    & 8.28\%  & 55.8   & 94.1\cr
        PWC-Net\_RVC\cite{sun2018pwc}       & -    & 4.90 & -    & 3.90 & -    & 11.63\% & 59.9   & 44.7\cr
        PRAFlow\_RVC(ours)                  & 1.34 & 3.56 & 0.77 & 2.48 & 1.13 & 5.43\%  & 64.0   & 25.6\cr
        \midrule
        IRR-PWC\cite{hur2019iterative}        & 2.51 & 4.58 & 1.92 & 3.84 & 1.63 & 7.65\% & - & -\cr
        ScopeFlow\cite{bar2020scopeflow}      & -    & 4.10 & -    & 3.59 & -    & 6.82\% & - & -\cr
        MaskFlowNet\cite{zhao2020maskflownet} & -    & 4.17 & -    & 2.52 & -    & 6.10\% & - & -\cr
        VCN\cite{yang2019volumetric}          & 2.24 & 4.40 & 1.66 & 2.81 & 1.16 & 6.30\% & - & -\cr
        RAFT\cite{teed2020raft}               & 1.27 & 2.86 & 0.77 & 1.61 & 0.63 & 5.10\% & - & -\cr
        \toprule
    \end{tabular}
    \end{threeparttable}
\end{table*}

\section{Conclusions}
We propose Pyramid Recurrent All-Pairs Flow (PRAFlow) for accurate high-resolution optical flow estimation, which is a pyramid coarse-to-fine version of RAFT\cite{teed2020raft}. Our method was submit to the optical flow task of the ECCV 2020 Robust Vision Challenge and got the 2nd place. The remaining experiments prove our method has strong cross dataset generalization. However, computing the correlation takes a lot of GPU memory even at quarter resolution. Therefore, our model occupies more computing resources when processing high-resolution images, so we had to downsample the resolution of input image. Next, we consider how to solve this problem to achieve more accurate optical flow estimation of high-resolution images.




\begin{appendices}
\section{Generalization ability}
The pretrain schedule aims to provide a good initialization for subsequent joint training schedule, and we evaluate the trained model on Sintel and KITTI datasets using their training data to validate the generalization performance of our model.
Validation results of model trained on FlyingChairs and FlyingThings3D datasets are shown in Table \ref{tab:things}, our method demonstrates good generalization performance across datasets.

\begin{table}[h]
    \caption{Validation Results of model trained on FlyingThings dataset}
    \label{tab:things}
    \centering
    \resizebox{0.5\textwidth}{!}{
        \begin{threeparttable}
        \begin{tabular}{ccccc}
            \toprule
            & Sintel \textit{Final} & Sintel \textit{Clean} & \multicolumn{2}{c}{KITTI} \cr
            \cmidrule(lr){2-3}\cmidrule(lr){4-5}
            Method        & AEPE     & AEPE    & AEPE    & F-all \cr
            \midrule
            PWC-Net\cite{sun2018pwc}               & 3.93      & 2.55      & 10.35  & 33.7     \cr
            MaskFlowNet\cite{zhao2020maskflownet}  & 3.72      & 2.33      & -      & -        \cr
            LiteFlowNet2\cite{hui2020lightweight}  & 3.78      & 2.24      & 8.97   & 25.9     \cr
            VCN\cite{yang2019volumetric}           & 3.68      & 2.21      & 8.36   & 25.1     \cr
            RAFT\cite{teed2020raft}                & 2.71      & 1.43      & 5.04   & 17.4     \cr
            PRAFlow\_RVC(ours)                     & 2.82      & 1.29      & 6.15   & 19.3     \cr
            \toprule
        \end{tabular}
        \end{threeparttable}
    }
\end{table}

\section{Good performance in small motion areas}
We found that PRAFlow has obvious advantages in estimating optical flow in small motion areas. Here we compare the s0-10, s10-40, and d0-10, d10-40 indicators of the MPI Sintel benchmark. Where d0-10 means endpoint error over regions closer than 10 pixels to the nearest occlusion boundary, s0-10 means endpoint error over regions with velocities lower than 10 pixels per frame. Detailed results as presented in Table \ref{tab:small-moving}.
\begin{table*}[h]
    \caption{Validation Results of different motion areas of Sintel dataset}
    \label{tab:small-moving}
    \centering
    \resizebox{0.8\textwidth}{!}{
        \begin{threeparttable}
        \begin{tabular}{ccccccccc}
        \toprule
        & \multicolumn{4}{c}{Sintel \textit{Final}} & \multicolumn{4}{c}{Sintel \textit{Clean}} \cr
        \cmidrule(lr){2-5}\cmidrule(lr){6-9}
        Method  & d0-10  & d10-60  & s0-10  & s10-60 & d0-10  & d10-60  & s0-10  & s10-60 \cr
        \midrule
        RAFT-TF\_RVC                          & 3.48  & 1.45  & 0.61   & 2.30    & 2.13  & 0.65  & 0.35  & 1.19 \cr
        C-RAFT\_RVC                           & 3.59  & 1.66  & 0.67   & 2.37    & 1.93  & 0.75  & 0.35  & 1.20 \cr
        VCN\_RVC\cite{yang2019volumetric}     & 4.13  & 1.92  & 0.83   & 2.99    & 2.86  & 0.92  & 0.51  & 1.48 \cr
        MaskFlowNet\cite{zhao2020maskflownet} & 3.78  & 1.75  & 0.59   & 2.39    & 2.74  & 0.91  & 0.36  & 1.29 \cr
        RAFT\cite{teed2020raft}               & 3.12  & 1.13  & 0.63   & 1.82    & 1.62  & 0.52  & 0.34  & 1.04 \cr
        PRAFlow\_RVC(ours)                    & 2.85  & 1.32  & 0.55   & 1.80    & 1.66  & 0.78  & 0.25  & 0.98 \cr
        \toprule
        \end{tabular}
        \end{threeparttable}
    }
\end{table*}

\section{Visualization}
\begin{figure*}[t] 
\centering 
\includegraphics[width=1\textwidth]{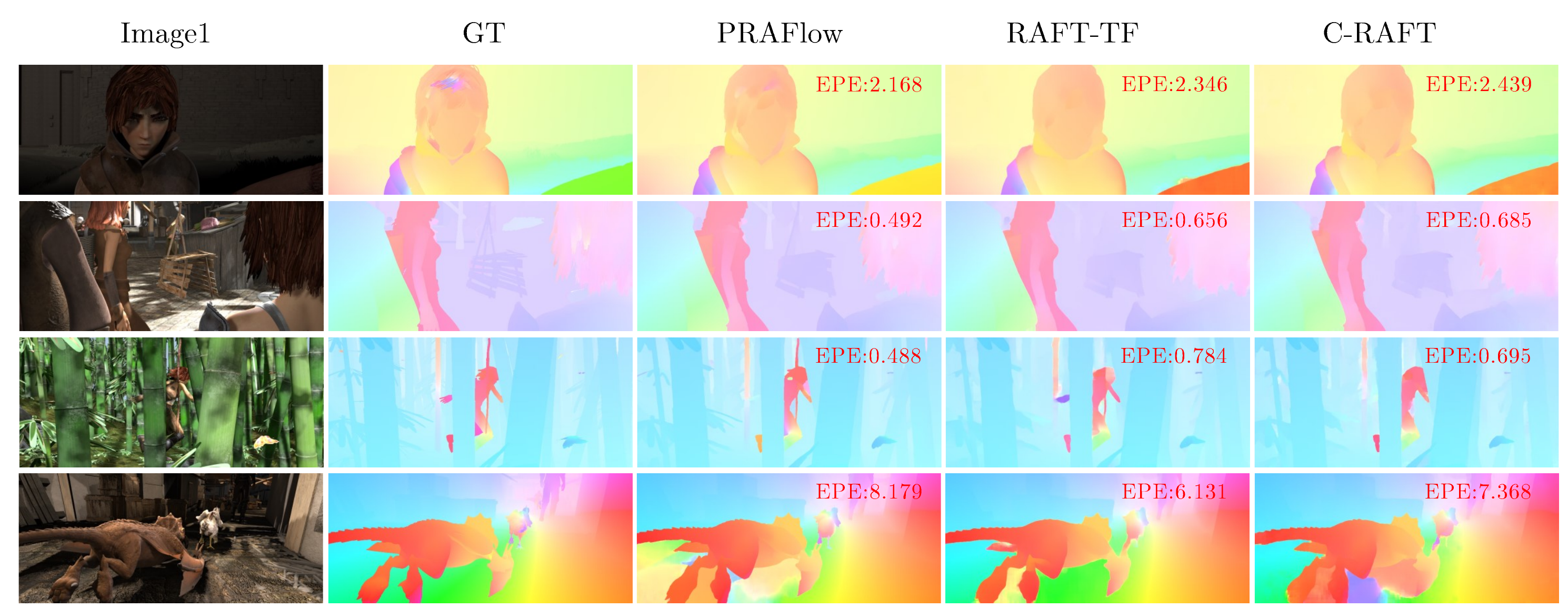}
\caption{\textbf{Visualization of predicted flow on Sintel Clean set}: (\textit{left to right}) input image1, ground truth flow, our PRAFlow\_RVC, RAFT-TF\_RVC and C-RAFT\_RVC} 
\label{fig-sintel}
\end{figure*}

\begin{figure*}[t] 
\centering 
\includegraphics[width=1\textwidth]{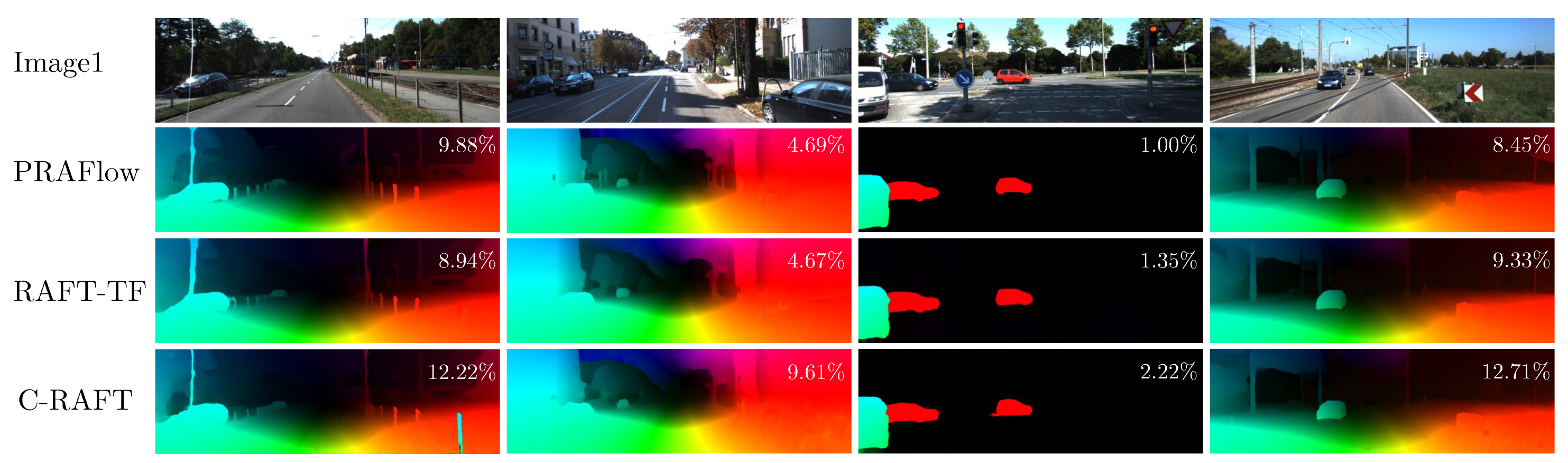}
\caption{\textbf{Visualization of predicted flow on KITTI dataset}: (\textit{top to down}) input image1, our PRAFlow\_RVC, RAFT-TF\_RVC and C-RAFT\_RVC} 
\label{fig-kitti}
\end{figure*}

The visualization results as comparisons are shown in Figure \ref{fig-sintel} and \ref{fig-kitti}. It shows that the details of the optical flow estimated by our method are better. 

\begin{figure*}[t] 
\centering 
\includegraphics[width=1\textwidth]{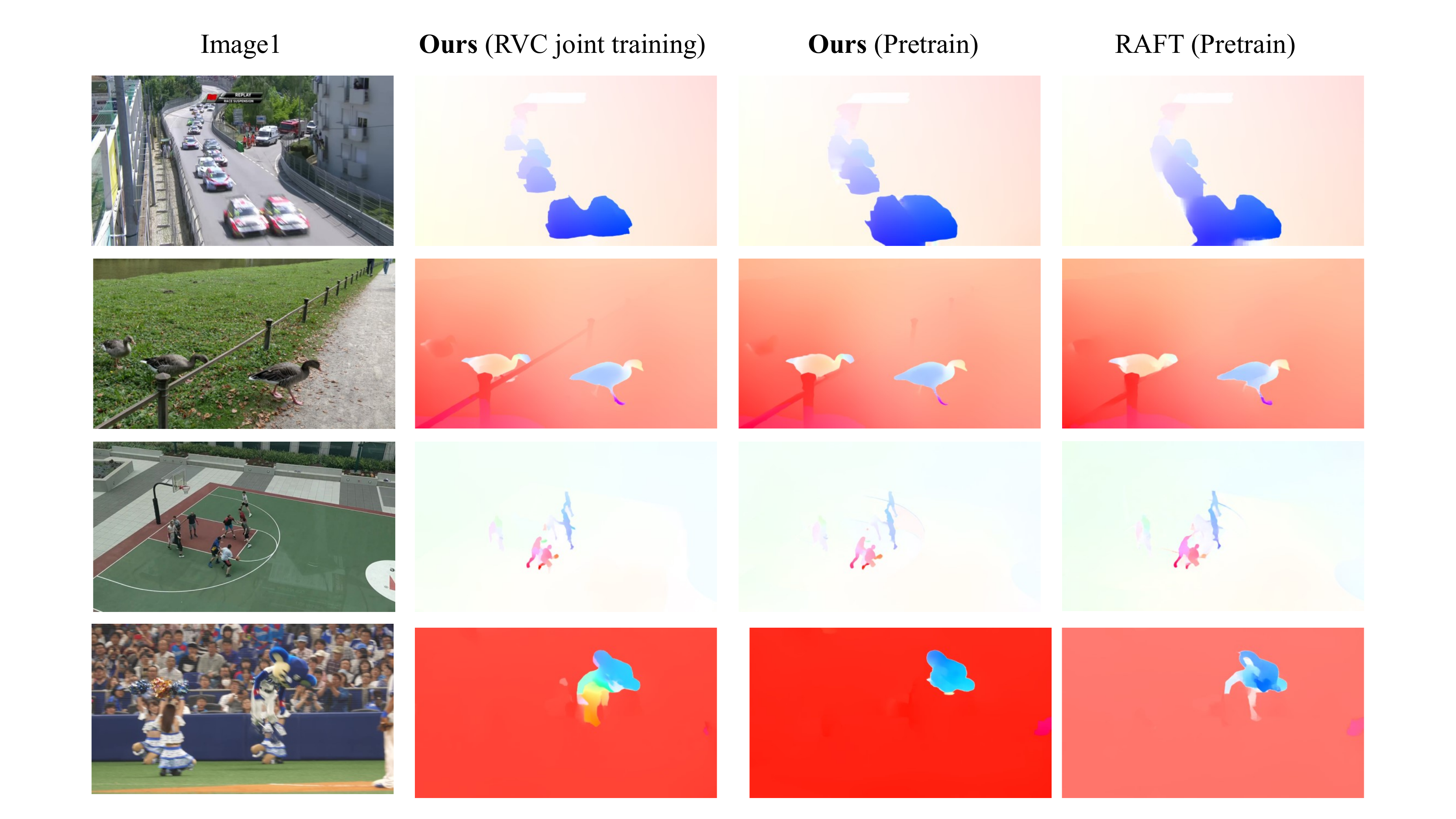}
\caption{\textbf{Visualization of predicted flow on DAVIS dataset}: (\textit{left to right}) input image1, our PRAFlow\_RVC(RVC joint training), our PRAFlow\_RVC(Pretrain) and RAFT(Pretrain)}  
\label{fig-davis}
\end{figure*}

We also test our model on DAVIS dataset, visualization results as comparisons are shown in Figure  \ref{fig-davis}. The visualization results show that our model obtained better results. Especially in the small motion area and high-resolution motion boundaries. But motion blur is still a challenge for both our model and RAFT.

\end{appendices}


\bibliographystyle{IEEEtran}
\bibliography{ref}
%



\end{document}